\documentclass[letterpaper, 10 pt, conference]{ieeeconf}%

\IEEEoverridecommandlockouts%

\usepackage{cite}
\usepackage{amsmath,amssymb,amsfonts}
\usepackage{graphicx}
\usepackage[table]{xcolor}
\usepackage{siunitx}
\usepackage{svg}
\usepackage{graphicx}
\usepackage{caption}
\usepackage{subcaption}
\usepackage{algorithm}
\usepackage{algorithmic}
\def\BibTeX{{\rm B\kern-.05em{\sc i\kern-.025em b}\kern-.08em
		T\kern-.1667em\lower.7ex\hbox{E}\kern-.125emX}}
\usepackage{afterpage}
\usepackage{bm}

\newcommand{\vc}[1]{\bm{#1}}

\newcommand{\matb}[1]{\begin{bmatrix}#1\end{bmatrix}}

\title{\LARGE \bf Control Barrier Functions for Safe Free-Flying Robotic Spacecraft Operations in Tumbling Target Capture}%

\author{Alexander Meinert$^{1}$, Peter Stadler$^{1}$, Niklas Baldauf$^{1}$ and Alen Turnwald$^{2}$%
\thanks{This work was supported by the Bavarian Ministry of Economic Affairs, Regional Development and Energy (grant no. MRF-2307-0009).}%
\thanks{$^{1}$Alexander Meinert, Peter Stadler and Niklas Baldauf are with the Space Applications Group, e:fs TechHub GmbH, Gaimersheim, Germany. {\tt\small \{Alexander.Meinert, Peter.Stadler, Niklas.Baldauf\}@efs-techhub.com}}
\thanks{$^{2}$Alen Turnwald is with the Faculty of Electrical Engineering and Information Technology, Ingolstadt University of Applied Sciences, Germany. {\tt\small Alen.Turnwald@thi.de}}%
}

\begin{document}

\maketitle
\thispagestyle{empty}
\pagestyle{empty}

\begin{abstract}
This paper presents a modular control barrier function (CBF) framework for safe free-flying robotic spacecraft operations during tumbling target capture. Motivated by latest ESA guidelines for safe close proximity operations, safety zones and requirements are translated into dedicated CBFs. The 13-DoF system is decomposed into translational, attitude, and robotic subsystems, each equipped with a safety filter that minimally modifies nominal control inputs in a lightweight quadratic program. The filters enforce a conical approach corridor, collision avoidance zone, attitude line-of-sight pointing, angular velocity limits, robotic joint limits, link-base collision avoidance, and actuator constraints. Dynamic coupling between subsystems is handled by treating upstream safe control commands as known interconnection inputs in the downstream safety filters, preserving modularity while supporting system-level safety. The framework is validated in an on-orbit servicing scenario, including final approach, angular rate synchronization, and tumbling target grasping, using the high-fidelity astrodynamics simulator Basilisk. Monte Carlo simulation results demonstrate runtime efficiency and operational safety for various tumbling rates.
\end{abstract}
\section{Introduction}
The growing number of satellites in Earth orbit and the associated collision risk create an increasing demand for on-orbit servicing (OOS) activities, such as active debris removal, inspection, and in-orbit maintenance. Robotic space manipulators offer a promising means to autonomously capture potentially uncooperative targets. However, close proximity operations (CPO) between two spacecraft are inherently safety-critical and require constraints and geometric safety zones to be considered throughout the mission timeline. To support the development of such missions, ESA has recently published guidelines for safe CPO \cite{ESA2024}, in line with existing requirements in the space industry \cite{ISO243302022}, and their application to mission concepts such as ClearSpace-1 is discussed in \cite{Vasconcelos2025}.

Constraint handling in this setting remains challenging. Incorporating operational constraints directly into Model Predictive Control (MPC) can lead to non-convex and computationally demanding optimal control problems, while Hamilton-Jacobi Reachability control design suffers from the curse of dimensionality for high-dimensional robotic spacecraft systems \cite{Wabersich2023}. Control Barrier Functions (CBFs), in contrast, provide a constructive way to define forward invariant safe sets by translating operational safety zones into control constraints. These constraints can be enforced through a one-step safety filter quadratic program (QP) with comparatively low computational cost \cite{Ames2019}. To this end, a CBF-QP safety layer can be combined with various nominal controllers, including unconstrained or learning-based approaches, since it minimally modifies the commanded input only when required for safety.
\begin{figure}[t!]
	\setlength{\belowcaptionskip}{-19pt}
	\centering
	\def\svgwidth{1.0\linewidth}
	\includegraphics[width=1.0\linewidth,page=1]{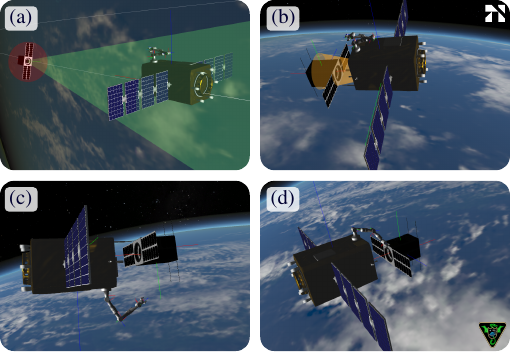}
	\caption{On-orbit servicing scenario sequence indicating selected operational safety zones. (a) Final approach (I) towards \textit{Go for Capture} decision point while maintaining the conical approach corridor (green) and spherical collision avoidance zone (red). (b) Attitude angular rate synchronization (II) while target pointing with rotational line-of-sight cone (yellow) towards grasping point. (c) Robotic grasping maneuver (III) during synchronized angular motion. (d) Target capture with the robotic end-effector.}
	\label{fig:vizard-scenario}
\end{figure}
Existing CBF applications in spacecraft and robotics typically focus on individual mission phases, subsystems, or constraint classes. For translational spacecraft relative motion, recent studies have investigated CBF formulations for rendezvous, docking corridors, and collision avoidance regions \cite{Breeden2022Docking, Meinert2026}. For the attitude subsystem, CBFs have been designed to constrain pointing maneuvers \cite{Tan2020} and angular velocity \cite{vanWijk2024}. In robotics, CBF-based filters have demonstrated safety guarantees for manipulators operating in cluttered environments \cite{Morton2025} and high-precision interaction tasks \cite{Shen2024}. Only a few works address CBF-based safety for robotic OOS missions, and these remain limited to a subset of constraints and degrees of freedom (DoF). Relevant studies include \cite{Loettgen2025, Sharifi2024} that assume a free-floating servicer base and primarily focus on CBFs for robotic joints and collision as well as the base angular velocity in \cite{Loettgen2025}, while attitude line-of-sight (LOS) and relative motion safety zones are not considered. In \cite{Shi2025}, the authors use a free-flying model of the space robot with a fully-actuated base, yet only employ link and base collision avoidance with the target and joint limit CBFs. None of the works cover tumbling target motion in their scenario setup.

In this paper, we develop a safe CBF-based control approach for the capture of an uncooperative tumbling target with a free-flying robotic spacecraft. The proposed framework addresses safety constraints across all 13 DoF and across the relevant OOS mission phases, from final approach to target synchronization and robotic grasping, while reflecting the geometric safety concepts defined in latest CPO guidelines.

In contrast to combined control strategies \cite{Invernizzi2023, Colmenarejo2018}, where a single controller is derived for the full-order robotic spacecraft system, we intentionally pursue a modular, subsystem-based control architecture. This reflects common space flight software development practice, where guidance, navigation, control, and robotic functions are often developed and validated as separate modules by different stakeholders. Such a modular structure is particularly relevant for OOS missions involving large consortia of hardware and software partners. It also allows subsystem controllers to be operated at different rates and to be activated or deactivated depending on the mission phase. At the same time, decoupled modeling improves runtime efficiency, which is important for resource-constrained space processors. The main technical challenge is then to guarantee safety despite the dynamic coupling between subsystems. To address this, we propose a safety layer consisting of dedicated safety filters for the translational, attitude, and robotic subsystems. The interdependence between these filters is explicitly considered by passing upstream safe commands to the downstream safety filters as known interconnection inputs. This preserves the computational advantages of modular control while making the subsystem QPs coupling-aware.

Our contributions are thus threefold:
\begin{itemize}
	\item A modular CBF-based safety framework for a 13-DoF free-flying robotic spacecraft, structured into translational, attitude, and robotic subsystem safety filters across final approach, target synchronization, and robotic grasping
	\item The integration of OOS safety zones and operational constraints consistent with ESA standards via dedicated CBF constraints for relative translational motion, attitude pointing and angular velocity, robotic joint limits, link-base collision avoidance, and input constraints
	\item The validation of the proposed approach in an on-orbit tumbling target capture scenario using the high-fidelity astrodynamics simulation framework \textit{Basilisk} \cite{Kenneally2020}
\end{itemize}
\section{Concept of Operations} \label{sec: Conops} 
The concept of operations for the considered OOS mission comprises three consecutive phases: (I) final approach, (II) angular rate synchronization, and (III) capture. Phases (I) and (II) correspond to the Close Rendezvous phase in the general-purpose ESA guidelines for safe CPO \cite{ESA2024}. In phase (I), the servicer approaches the target along the target angular momentum vector inside a conical Approach Corridor (AC) while entering the spherical Keep-Out Zone (KOZ). The KOZ denotes the operational region in which collisions can occur, since its radius is defined by the sum of the largest dimensions of the servicer and target clearance envelopes. Accordingly, 6-DoF relative navigation and closed-loop control are mandatory inside the KOZ and enforced by the bounds of the AC. The final approach starts at the \textit{Go for KOZ} decision point and ends at the \textit{Go for Capture} decision point. We further introduce a mission-specific spherical Collision Avoidance Zone (CAZ), which defines the minimum distance between servicer and target to be maintained at all times. In phase (II), the servicer synchronizes its angular rate with the tumbling target, while aligning its boresight direction with the grasping point. Phase (III) starts with the grasping maneuver of the robotic manipulator and ends with the successful capture of the target once the end-effector (EE) reaches the grasping point. Figure \ref{fig:vizard-scenario} illustrates the OOS mission phases and selected operational safety zones.
\section{Problem Statement}\label{sec: Problem Statement}
The free-flying servicing spacecraft equipped with a 7-DoF robotic manipulator operating in close proximity of a tumbling client has the full-order control input
\begin{equation}
	\vc{u} = \begin{bmatrix} \vc{f}_B^\top & \vc{\tau}_B^\top & \vc{\tau}_M^\top \end{bmatrix}^\top \in \mathbb{R}^{13},
\end{equation}
where $\vc{f}_B\in\mathbb{R}^3$ denotes the spacecraft translational force, $\vc{\tau}_B\in\mathbb{R}^3$ is the spacecraft attitude torque, and $\vc{\tau}_M\in\mathbb{R}^7$ are the manipulator joint torques.
For the subsystem-based approach, the translational, attitude, and robotic dynamics are modeled as follows.
\subsection{Translational Subsystem}
Considering a circular target orbit, the relative translational motion is given by the Clohessy-Wiltshire-Hill equations as
\begin{equation}\label{eq: rel motion system}
	\small
	\setlength{\arraycolsep}{2.5pt}
	\renewcommand{\arraystretch}{1.0}
	\begin{bmatrix}
		\dot{r}_1 \\ \dot{r}_2 \\ \dot{r}_3 \\ \dot{v}_1 \\ \dot{v}_2 \\ \dot{v}_3
	\end{bmatrix} = 
	\begin{bmatrix}
		0 & 0 & 0 & 1 & 0 & 0 \\
		0 & 0 & 0 & 0 & 1 & 0 \\
		0 & 0 & 0 & 0 & 0 & 1 \\
		3n^2 & 0 & 0 & 0 & 2n & 0 \\
		0 & 0 & 0 & -2n & 0 & 0 \\
		0 & 0 & -n^2 & 0 & 0 & 0 
	\end{bmatrix}
	\begin{bmatrix}
		r_1 \\ r_2 \\ r_3 \\ v_1 \\ v_2 \\ v_3
	\end{bmatrix}
	+
	\frac{1}{m} 
	\begin{bmatrix}
		0 & 0 & 0 \\
		0 & 0 & 0 \\
		0 & 0 & 0 \\
		1 & 0 & 0 \\
		0 & 1 & 0 \\
		0 & 0 & 1
	\end{bmatrix}
	\begin{bmatrix}
		f_1 \\ f_2 \\ f_3
	\end{bmatrix}
	\normalsize
\end{equation}
in which $\vc{x}_{\mathrm{tr}}=[\vc{r}_H^\top\;\vc{v}_H^\top]^\top$, $\vc{r}_H\in\mathbb{R}^3$ denotes the relative position of the servicer with respect to the target in the Hill frame $\mathcal{F}_H$, and $\vc{v}_H=\dot{\vc{r}}_H\in\mathbb{R}^3$ is the corresponding relative velocity. The mass and mean orbital motion are expressed by the parameters $m$ and $n$, respectively, while $\vc{f}_H=[f_1\;f_2\;f_3]^\top$ is the actuated input force. The three components of the state and input represent the R-bar, V-bar, and H-bar direction given in the target-centered Hill frame.
\subsection{Attitude Subsystem}
Let the quaternion $\vc{q}=[q_0,q_1,q_2,q_3]^\top\in\mathbb{S}^3$ define the servicing spacecraft attitude, with $q_0$ denoting the scalar part, and let $\vc{\omega}\in\mathbb{R}^3$ be its angular velocity expressed in the servicer body frame $\mathcal{F}_B$. Then the attitude kinematics and dynamics are formulated as
\begin{equation}\label{eq: Attitude kinematics}
	\dot{\vc{q}} = \frac{1}{2}\,\vc{S}_q(\vc{q})\,\vc{\omega}\,,\quad \vc{S}_q(\vc{q}) = \matb{-q_1 & -q_2 & -q_3 \\ q_0 & -q_3 & q_2 \\ q_3 & q_0 & -q_1 \\ -q_2 & q_1 & q_0}
\end{equation}
\begin{equation}\label{eq: Attitude dynamics}
	\vc{I}\,\dot{\vc{\omega}} = \vc{\tau}_B - \vc{\omega} \times (\vc{I}\,\vc{\omega}),
\end{equation}
where $\vc{I}$ is the inertia of the servicer.
\subsection{Robotic Subsystem}
The robotic manipulator subsystem is modeled relative to the servicer base frame by
\begin{equation}
	\vc{M}(\vc{\theta})\ddot{\vc{\theta}} +	\vc{C}(\vc{\theta},\dot{\vc{\theta}})\dot{\vc{\theta}} = \vc{\tau}_M,
\end{equation}
with the mass matrix $\vc{M}$, the Coriolis and centrifugal forces $\vc{C}$, the joint angles $\vc{\theta}\in\mathbb{R}^7$, the joint velocities $\dot{\vc{\theta}}\in\mathbb{R}^7$, and its accelerations $\ddot{\vc{\theta}}\in\mathbb{R}^7$. The position and twist in the base frame are obtained from forward kinematics and the geometric Jacobian $\vc{J}_E$,
\begin{equation}
	\vc{p}_{E,B}=\vc{f}_{\mathrm{FK}}(\vc{\theta}),\qquad
	\dot{\vc{p}}_{E,B}=\vc{J}_E(\vc{\theta})\dot{\vc{\theta}},
\end{equation}
here derived for the end-effector mapped to the operational space.
\subsection{Reference Frame Convention} \label{subsec:reference_frames}
While the controller and CBF design are based on different task-dependent frames internally, as outlined in the previous subsystem modeling, a frame convention is required for thorough analysis of the servicing scenario. We define a target-centered non-rotating frame, hereinafter called the \textit{reference} frame $\mathcal{F}_R$, whose origin is attached to the target center of mass and whose attitude is inertially fixed to the initial servicer attitude; hence, $\mathcal{F}_R$ does not rotate with the tumbling target.
\section{Method} \label{sec: Method}%
\begin{figure*}[t!]
	\setlength{\belowcaptionskip}{-15pt}
	\centering
	\includegraphics[width=0.8\linewidth,page=2]{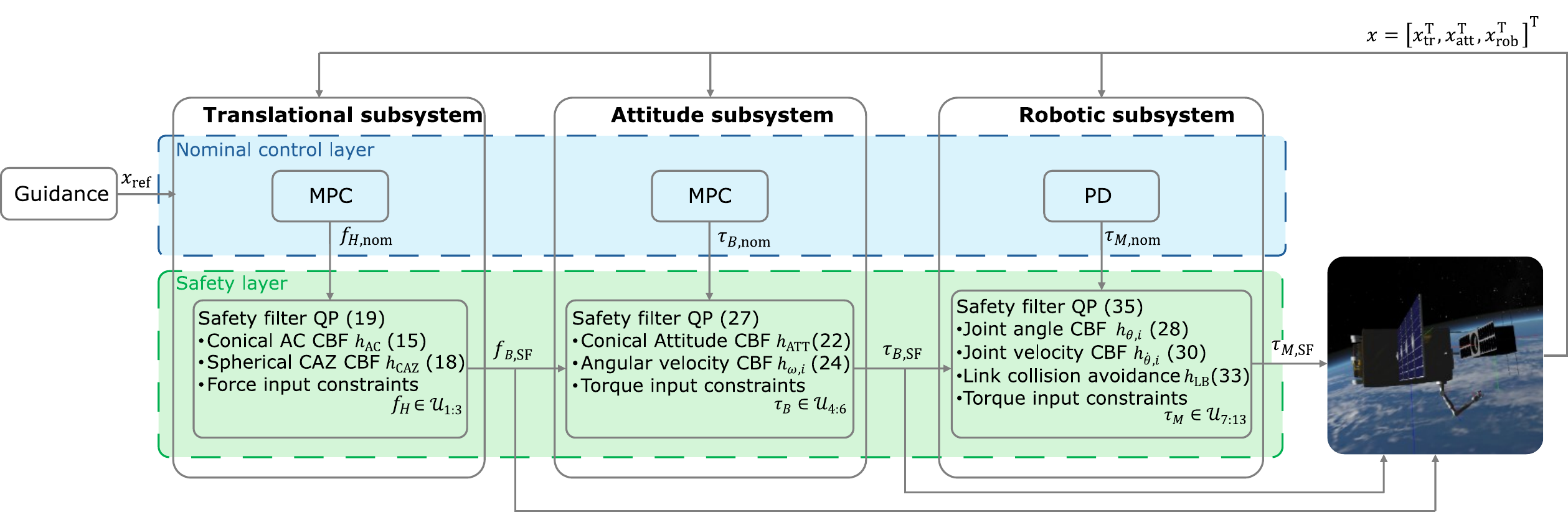}
	\caption{Safe modular control architecture with nominal subsystem controllers and CBF-QP safety filters for translational, attitude, and robotic constraint enforcement. The filters minimally modify the nominal commands while accounting for subsystem coupling through causally updated upstream safe inputs.}
	\label{fig:control_safety_framework}
\end{figure*}

For the nonlinear control-affine system
\begin{equation}\label{eq:control-affine system}
	\dot{\vc{x}}
	=
	\tilde{\vc{f}}(\vc{x},t,\vc{\xi})
	+
	\vc{g}(\vc{x},t)\vc{u},
	\qquad
	\tilde{\vc{f}}(\vc{x},t,\vc{\xi})
	=
	\vc{f}(\vc{x},t)+\vc{\xi}(t),
\end{equation}
where $\vc{x} \in \mathcal{X} \subset \mathbb{R}^n$ and
$\vc{u} \in \mathcal{U} \subset \mathbb{R}^m$ denote the state and control input, respectively, $\vc{\xi}$ is introduced as a known exogenous coupling term that appears in the drift dynamics $\tilde{\vc{f}}(\vc{x},t,\vc{\xi})$.
The safe set is defined as the zero-superlevel set
\begin{equation}
	\mathcal{S}(t)
	=
	\{\vc{x} \in \mathbb{R}^n \mid h(\vc{x},t) \geq 0\}
\end{equation}
of a continuously differentiable function $h(\vc{x},t)$. According to \cite{Ames2019}, if there exists an extended class-$\mathcal{K}$ function $\alpha$ such that the safety condition
\begin{equation} \label{eq:CBF rel deg 1}
	\small
	\sup\limits_{\vc{u} \in \mathcal{U}}
	\left[
	\frac{\partial h}{\partial t}
	+
	L_{\tilde f}h(\vc{x},t,\vc{\xi})
	+
	L_gh(\vc{x},t)\vc{u}
	\right]
	\geq
	-\alpha(h(\vc{x},t))
\end{equation}
can be satisfied for all $\vc{x}\in\mathcal{S}(t)$, then any locally Lipschitz controller satisfying \eqref{eq:CBF rel deg 1} renders $\mathcal{S}(t)$ forward invariant, and $h(\vc{x},t)$ is said to be a Control Barrier Function. Following the results from \cite{Breeden2023Robust}, this formulation can be extended to position-level safety constraints $h(\vc{x},t)$ with relative degree two, where the control input appears in the second total time derivative along the system dynamics. For this purpose, the auxiliary barrier function
\begin{equation}
	H(\vc{x},t,\vc{\xi})
	=
	h(\vc{x},t)
	+
	\frac{|\dot{h}(\vc{x},t,\vc{\xi})|\dot{h}(\vc{x},t,\vc{\xi})}{2u_{\max}}
\end{equation}
is introduced with
\begin{equation}
	\dot{H}(\vc{x},\vc{u},t,\vc{\xi})
	=
	\dot{h}(\vc{x},t,\vc{\xi})
	+
	\frac{
		|\dot{h}(\vc{x},t,\vc{\xi})|
		\ddot{h}(\vc{x},\vc{u},t,\vc{\xi})
	}{u_{\max}} .
\end{equation}
The corresponding CBF condition with relative degree two is given by
\begin{equation} \label{eq:CBF rel deg 2}
	\small
	\begin{aligned}
		\Phi_h(\vc{x},\vc{u},t,\vc{\xi})
		={}&
		\frac{\partial H}{\partial t}
		+ L_{\tilde f} H
		+ L_g H\,\vc{u}
		+ \alpha(H)
		- \epsilon_h ,
	\end{aligned}
\end{equation}
where all terms involving $H$ are evaluated at $(\vc{x},t,\vc{\xi})$, $\epsilon_h$ accounts for discretization effects by creating an offset to the safe set boundary, and $u_{\max}$ is set equal to the maximal control authority regarding the respective subsystem. Safety is enforced by requiring
\begin{equation}
	\sup\limits_{\vc{u} \in \mathcal{U}}
	\Phi_h(\vc{x},\vc{u},t,\vc{\xi})
	\geq 0 .
\end{equation}
We further remark that $\gamma_h$ is introduced as a tuning parameter that lower bounds the decay rate of the CBF value function such that $\alpha(H(\vc{x},t))=\gamma_h H(\vc{x},t)$. Figure \ref{fig:control_safety_framework} depicts the overall control architecture, where we assume a causal update order from translational to attitude and robotic subsystems, with downstream filters running at least as fast as the upstream signal.
Note that in contrast to robust CBF formulations for unknown bounded disturbances, no worst-case maximization or additional disturbance margin is introduced, since the upstream coupling input $\vc{\xi}$ is assumed to be known when the corresponding safety filter is evaluated.
\subsection{CBFs for the Translational Subsystem}
\subsubsection{Conical Approach Corridor Keep-In Zone}
The approach corridor axis is defined as a constant unit vector $\vc{d}_R$ in the reference frame $\mathcal{F}_R$. Since the conical CBF is designed for the translational relative motion expressed in the Hill frame $\mathcal{F}_H$, this axis becomes time-varying as
\begin{equation}
	\vc{d}_H(t)=\vc{R}_{HR}(t)\vc{d}_R ,
\end{equation}
considering the rotation matrix $\vc{R}_{HR}(t)\in SO(3)$ from $\mathcal{F}_R$ to $\mathcal{F}_H$. We define the conical AC CBF as
\begin{equation}\label{eq:CBF-AC}
	h_{\mathrm{AC}}(\vc{r}_H,t)
	=
	\vc{d}_H^\top(t)\vc{r}_H
	-
	\|\vc{r}_H\|\cos(\psi_{\mathrm{AC}})
	\geq 0 ,
\end{equation}
where $\psi_{\mathrm{AC}}$ is the half-cone angle of the admissible corridor. This yields its first derivative
\begin{equation}
	\dot h_{\mathrm{AC}}
	=
	\left(
	\vc{d}_H
	-
	\frac{\vc{r}_H}{\|\vc{r}_H\|}\cos(\psi_{\mathrm{AC}})
	\right)^\top
	\vc{v}_H
	+
	\dot{\vc{d}}_H^\top \vc{r}_H ,
\end{equation}
while the time variation of the corridor induced by the frame transformation enters the CBF derivative explicitly through $\dot{\vc{d}}_H$. Accordingly, the second derivative can be written in control-affine form as
\begin{equation}
	\ddot h_{\mathrm{AC}}
	=
	L_{\tilde f}^2h_{\mathrm{AC}}
	+
	L_gL_{\tilde f}h_{\mathrm{AC}}\vc{f}_H ,
\end{equation}
where $L_{\tilde f}^2h_{\mathrm{AC}}$ collects all drift terms, including the time-varying terms induced by $\dot{\vc{d}}_H$ and $\ddot{\vc{d}}_H$, while $L_gL_{\tilde f}h_{\mathrm{AC}}\vc{f}_H$ contains the affine dependence on the control input force $\vc{f}_H$. Since the translational filter is evaluated first, no upstream input is considered and $\vc{\xi}_{\mathrm{tr}}=\vc{0}$. For brevity, the detailed substitutions are omitted for the remaining CBFs, but the same differentiation procedure is applied to each constraint.
\subsubsection{Spherical Base Collision Avoidance Zone}
The servicer is required to preserve a minimum distance to the target when approaching the \textit{Go for Capture} decision point. This requirement is translated into a spherical CAZ around the target and encoded as the safety constraint
\begin{equation}\label{eq:CBF-CAZ}
	h_{\mathrm{CAZ}}(\vc{r}_H)
	=
	\|\vc{r}_H\|^2
	-
	R_{\mathrm{CAZ}}^2
	\geq 0 ,
\end{equation}
using the relative position $\vc{r}_H$ and the minimum safety distance $R_{\mathrm{CAZ}}$.
Substituting the CBFs \eqref{eq:CBF-AC} and \eqref{eq:CBF-CAZ} into \eqref{eq:CBF rel deg 2}, the translational safety filter QP, repeatedly solved at every time step, yields
\begin{equation}\label{eq:translation_qp}
	\begin{aligned}
		\vc{f}_{H,\text{SF}}=\arg\min_{\vc{f}_H}\quad&\|\vc{f}_H-\vc{f}_{H,\text{nom}}\|^2\\
		\mathrm{s.t.}\quad&\Phi_{\mathrm{AC}}(\vc{x}_{\mathrm{tr}},\vc{f}_H,t)\geq0\\
		&\Phi_{\mathrm{CAZ}}(\vc{x}_{\mathrm{tr}},\vc{f}_H)\geq0\\
		&\vc{f}_H\in\mathcal U_{\mathrm{1:3}} .
	\end{aligned}
\end{equation}
The safety filter minimally adjusts the nominal control input force $\vc{f}_{H,\text{nom}}$ such that the conical AC and spherical CAZ as well as the input constraints are maintained. Since all constraints are affine in the decision variable, \eqref{eq:translation_qp} provides a runtime-efficient implementation suitable for online deployment on resource-constrained processors \cite{Meinert2026}. To this end, the resulting safe input $\vc{f}_{H,\text{SF}}$ is transformed to the servicer body frame and represents the actuated force $\vc{f}_{B,\text{SF}}$.
\subsection{CBFs for the Attitude Subsystem}
\subsubsection{Time-Varying Attitude Cone Keep-In Zone}
We introduce a LOS pointing constraint to keep the target within the admissible sensor cone. The resulting attitude keep-in zone bounds the relative attitude during operations inside the KOZ, thereby addressing collision avoidance while also serving the practical motivation to maintain camera-based perception of the grasping point, for which an EE-mounted camera alone would be insufficient in close proximity. This constraint is consistent with the ESA 6-DoF relative navigation requirement, but is typically neglected in the aforementioned CBF-based OOS literature. Let $\vc{d}_B$ denote the constant unit boresight vector expressed in the servicer body frame $\mathcal{F}_B$. The LOS direction from the servicer base to the target grasping point $\vc{p}_{G,T}$ is defined in the reference frame $\mathcal{F}_R$ as
\begin{equation}
	\vc{s}_R(t)
	=
	\frac{\vc{R}_{RT}(t)\vc{p}_{G,T} - \vc{r}_R(t)}{\|\vc{R}_{RT}(t)\vc{p}_{G,T} - \vc{r}_R(t)\|},
\end{equation}
where $\vc{r}_R$ is the relative position of the servicer with respect to the target center and $\vc{R}_{RT}(t)\in SO(3)$ rotates the target-fixed grasping point from the tumbling target frame $\mathcal{F}_T$ to $\mathcal{F}_R$. Using the attitude quaternion $\vc{q}$, the LOS direction is transformed into the servicer body frame by
\begin{equation}
	\vc{s}_B(\vc{q},t)
	=
	\vc{R}_{BR}(\vc{q})\vc{s}_R(t),
\end{equation}
where $\vc{R}_{BR}(\vc{q})\in SO(3)$ denotes the rotation matrix from $\mathcal{F}_R$ to $\mathcal{F}_B$. Thus, the attitude cone CBF is given by
\begin{equation}\label{eq:CBF-ATT}
	h_{\mathrm{ATT}}(\vc{q},t)
	=
	\vc{d}_B^\top\vc{s}_B(\vc{q},t)
	-
	\cos(\beta_{\mathrm{ATT}})
	\geq 0 ,
\end{equation}
where $\beta_{\mathrm{ATT}}$ is the half-cone angle of the admissible pointing cone. The time variation of $h_{\mathrm{ATT}}$ is induced by both the changing LOS vector $\vc{s}_R(t)$ and the servicer attitude motion. Since the attitude torque $\vc{\tau}_B$ enters the second time derivative of \eqref{eq:CBF-ATT}, the corresponding CBF constraint is written as
\begin{equation}
	\Phi_{\mathrm{ATT}}(\vc{x}_{\mathrm{att}},\vc{\tau}_B,t,\vc{\xi}_{\mathrm{att}})\geq0 ,
\end{equation}
with $\vc{x}_{\mathrm{att}}=[\vc{q}^\top\;\vc{\omega}^\top]^\top$ and let $\vc{\xi}_{\mathrm{att}}$ be the coupling term induced by the translational input $\vc{f}_{B,\mathrm{SF}}$.
\subsubsection{Servicer Angular Velocity Limits}
The angular velocity of the servicer is constrained component-wise in the body frame. For each axis $i\in\{1,2,3\}$, the upper and lower angular velocity limits are encoded as
\begin{equation}\label{eq:CBF-omega}
	h_{\omega,i}^{+}(\vc{\omega})
	=
	\omega_{i,\max}-\omega_i
	\geq0,
	\quad
	h_{\omega,i}^{-}(\vc{\omega})
	=
	\omega_i-\omega_{i,\min}
	\geq0 .
\end{equation}
\begin{figure*}[t]
	\setlength{\belowcaptionskip}{-16pt}
	\centering
	\includegraphics[width=1.0\linewidth,page=3]{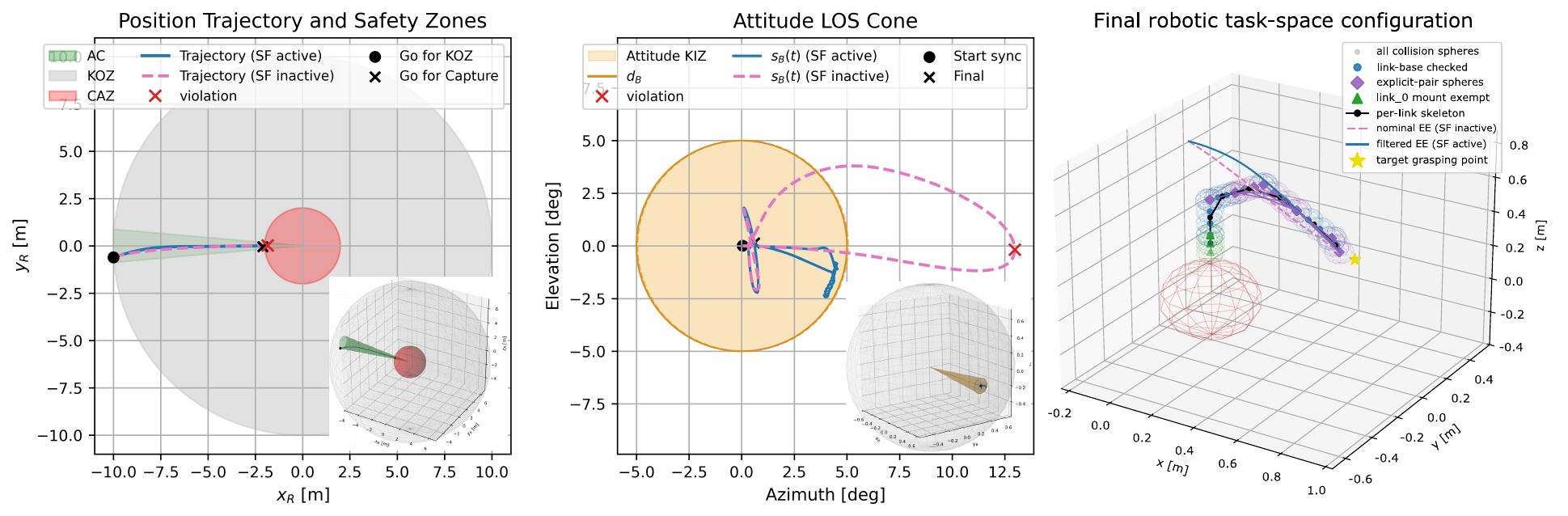}
	\caption{Translational motion profile of the servicer position including operational safety zones expressed in $\mathcal{F}_R$ (left), attitude trajectory regarding the LOS cone in $\mathcal{F}_B$ (center), and robotic EE trajectory in task space $\mathcal{F}_B$ (right). Both nominal (dashed) and filtered (solid) trajectories are illustrated.}
	\label{fig:results_trajectories}
\end{figure*}
As the attitude torque appears in the first derivative of $\vc{\omega}$ through \eqref{eq: Attitude dynamics}, these constraints are enforced with relative-degree-one CBF conditions
\begin{equation}
	\small
	\varphi_{\omega,i}^{+}(\vc{x}_{\mathrm{att}},\vc{\tau}_B,\vc{\xi}_{\mathrm{att}})
	=
	\dot h_{\omega,i}^{+}(\vc{x}_{\mathrm{att}},\vc{\tau}_B,\vc{\xi}_{\mathrm{att}})
	+
	\gamma_{\omega}^{+}h_{\omega,i}^{+}(\vc{\omega})
	-
	\epsilon_{\omega}^{+}
	\geq0 ,
\end{equation}
\begin{equation}
	\small
	\varphi_{\omega,i}^{-}(\vc{x}_{\mathrm{att}},\vc{\tau}_B,\vc{\xi}_{\mathrm{att}})
	=
	\dot h_{\omega,i}^{-}(\vc{x}_{\mathrm{att}},\vc{\tau}_B,\vc{\xi}_{\mathrm{att}})
	+
	\gamma_{\omega}^{-}h_{\omega,i}^{-}(\vc{\omega})
	-
	\epsilon_{\omega}^{-}
	\geq0 .
\end{equation}
Based on the CBFs \eqref{eq:CBF-ATT} and \eqref{eq:CBF-omega}, and using the interconnection input $\vc{\xi}_{\mathrm{att}}$, the attitude safety filter QP results in
\begin{equation}\label{eq:attitude_qp}
	\begin{aligned}
		\vc{\tau}_{B,\text{SF}}=\arg\min_{\vc{\tau}_B}\quad&\|\vc{\tau}_B-\vc{\tau}_{B,\text{nom}}\|^2\\
		\mathrm{s.t.}\quad&\Phi_{\mathrm{ATT}}(\vc{x}_{\mathrm{att}},\vc{\tau}_B,t,\vc{\xi}_{\mathrm{att}})\geq0\\
		&\varphi_{\omega,i}^{+}(\vc{x}_{\mathrm{att}},\vc{\tau}_B,\vc{\xi}_{\mathrm{att}})\geq0,\quad i=1,2,3\\
		&\varphi_{\omega,i}^{-}(\vc{x}_{\mathrm{att}},\vc{\tau}_B,\vc{\xi}_{\mathrm{att}})\geq0,\quad i=1,2,3\\
		&\vc{\tau}_B\in\mathcal U_{\mathrm{4:6}} .
	\end{aligned}
\end{equation}
incorporating the quadratic cost function that minimizes the deviation of the modified control input from the nominal actuation torque. In this regard, the attitude cone KIZ, the angular velocity limits, and the input constraints are satisfied. The safe control input torque $\vc{\tau}_{B,\text{SF}}$ is expressed in $\mathcal{F}_B$ and then applied to the system.
\subsection{CBFs for the Robotic Subsystem}
\subsubsection{Robotic Joint Angle Limits}
The manipulator joint angles are constrained component-wise to remain inside their admissible ranges. For each joint $i\in\{1,\dots,7\}$, the joint angle limit is encoded as
\begin{equation}\label{eq:CBF-joint-angle}
	h_{\theta,i}(\vc{\theta})
	=
	(\theta_i-\theta_{i,\min})
	(\theta_{i,\max}-\theta_i)
	\geq0 .
\end{equation}
As the joint torque $\vc{\tau}_M$ enters the second derivative of \eqref{eq:CBF-joint-angle}, the corresponding CBF constraint is written as
\begin{equation}
	\Phi_{\theta,i}(\vc{x}_{\mathrm{rob}},\vc{\tau}_M,\vc{\xi}_{\mathrm{rob}})\geq0,
	\qquad i=1,\dots,7 ,
\end{equation}
with $\vc{x}_{\mathrm{rob}}=[\vc{\theta}^\top\;\dot{\vc{\theta}}^\top]^\top$ and the coupling input $\vc{\xi}_{\mathrm{rob}}$ induced by the upstream safe force and attitude torque commands.
\subsubsection{Robotic Joint Velocity Limits}
Analogously, the manipulator joint velocities are constrained component-wise. For each joint velocity $i\in\{1,\dots,7\}$, the upper and lower velocity limits are given as
\begin{equation}\label{eq:CBF-joint-velocity}
	h_{\dot{\theta},i}^{+}(\dot{\vc{\theta}})
	=
	\dot{\theta}_{i,\max}-\dot{\theta}_i
	\geq0,
	\qquad
	h_{\dot{\theta},i}^{-}(\dot{\vc{\theta}})
	=
	\dot{\theta}_i-\dot{\theta}_{i,\min}
	\geq0 .
\end{equation}
Since the joint torque $\vc{\tau}_M$ appears in the first derivative of $\dot{\vc{\theta}}$ through the robotic dynamics, these constraints are enforced with relative-degree-one CBF conditions
\begin{equation}
	\small
	\varphi_{\dot{\theta},i}^{+}(\vc{x}_{\mathrm{rob}},\vc{\tau}_M,\vc{\xi}_{\mathrm{rob}})
	=
	\dot h_{\dot{\theta},i}^{+}(\vc{x}_{\mathrm{rob}},\vc{\tau}_M,\vc{\xi}_{\mathrm{rob}})
	+
	\gamma_{\dot{\theta}}^{+}h_{\dot{\theta},i}^{+}(\dot{\vc{\theta}})
	-
	\epsilon_{\dot{\theta}}^{+}
	\geq0 ,
\end{equation}
\begin{equation}
	\small
	\varphi_{\dot{\theta},i}^{-}(\vc{x}_{\mathrm{rob}},\vc{\tau}_M,\vc{\xi}_{\mathrm{rob}})
	=
	\dot h_{\dot{\theta},i}^{-}(\vc{x}_{\mathrm{rob}},\vc{\tau}_M,\vc{\xi}_{\mathrm{rob}})
	+
	\gamma_{\dot{\theta}}^{-}h_{\dot{\theta},i}^{-}(\dot{\vc{\theta}})
	-
	\epsilon_{\dot{\theta}}^{-}
	\geq0 .
\end{equation}
\subsubsection{Spherical Link-Base Collision Avoidance}
To avoid collisions between the manipulator and the servicer base, the robotic links are approximated by collision spheres. For each link, three points are used to cover the arm geometry. All quantities in this constraint are expressed in the servicer base frame $\mathcal{F}_B$. Let $\vc{p}_i(\vc{\theta})$ denote the center of the $i$-th robotic collision sphere and let $\rho_i$ be its radius. Furthermore, let $\vc{c}_{\mathrm{base}}$ and $R_{\mathrm{base}}$ denote the center and radius of the spherical base collision zone. The corresponding spherical link-base CBF is defined as
\begin{equation}\label{eq:CBF-link-base}
	h_{\mathrm{LB},i}(\vc{\theta})
	=
	\|\vc{p}_i(\vc{\theta})-\vc{c}_{\mathrm{base}}\|^2
	-
	(\rho_i+R_{\mathrm{base}})^2
	\geq0 .
\end{equation}
While this sphere-based approximation provides a rather conservative collision model, other geometric primitives, such as capsules or ellipsoids, can be incorporated in the same framework and may reduce conservativeness for specific link and base geometries, with the possibility of increasing the computational cost. Since the joint torque $\vc{\tau}_M$ enters the second derivative of \eqref{eq:CBF-link-base}, the corresponding CBF constraint is written as
\begin{equation}
	\Phi_{\mathrm{LB},i}(\vc{x}_{\mathrm{rob}},\vc{\tau}_M,\vc{\xi}_{\mathrm{rob}})\geq0 ,
\end{equation}
for all considered robotic collision spheres $i\in\mathcal{P}_{\mathrm{LB}}$. Taking into account the CBFs \eqref{eq:CBF-joint-angle}, \eqref{eq:CBF-joint-velocity}, and \eqref{eq:CBF-link-base}, the resulting safety filter QP for the robotic subsystem is given by
\begin{equation}\label{eq:robotic_qp}
	\begin{aligned}
		\vc{\tau}_{M,\text{SF}}=\arg\min_{\vc{\tau}_M}\quad&\|\vc{\tau}_M-\vc{\tau}_{M,\text{nom}}\|^2\\
		\mathrm{s.t.}\quad&\Phi_{\theta,i}(\vc{x}_{\mathrm{rob}},\vc{\tau}_M,\vc{\xi}_{\mathrm{rob}})\geq0,\quad i=1,\dots,7\\
		&\varphi_{\dot{\theta},i}^{+}(\vc{x}_{\mathrm{rob}},\vc{\tau}_M,\vc{\xi}_{\mathrm{rob}})\geq0,\quad i=1,\dots,7\\
		&\varphi_{\dot{\theta},i}^{-}(\vc{x}_{\mathrm{rob}},\vc{\tau}_M,\vc{\xi}_{\mathrm{rob}})\geq0,\quad i=1,\dots,7\\
		&\Phi_{\mathrm{LB},i}(\vc{x}_{\mathrm{rob}},\vc{\tau}_M,\vc{\xi}_{\mathrm{rob}})\geq0,\quad i\in\mathcal{P}_{\mathrm{LB}}\\
		&\vc{\tau}_M\in\mathcal U_{\mathrm{7:13}} .
	\end{aligned}
\end{equation}
\section{On-Orbit Servicing Simulation Study} \label{sec: Numerical Results}%

\subsection{Nominal Controllers}
For the nominal controllers, we use an MPC for both translational and attitude control as derived and parameterized in \cite{Meinert2026, Stadler2026}, respectively, as well as a joint space PD computed torque controller \cite{Morton2025}. The interested reader is referred to the references for detailed derivations. However, we emphasize that any nominal control signal is valid for usage in combination with the safety filters. Regarding the MPC, all constraints are inactive in the optimization problem to eventually obtain unsafe nominal control inputs and demonstrate the constraint satisfaction by the safety layer.
\begin{figure}[t!]
	\setlength{\belowcaptionskip}{-15pt}
	\centering
	\includegraphics[width=1.0\linewidth,page=4]{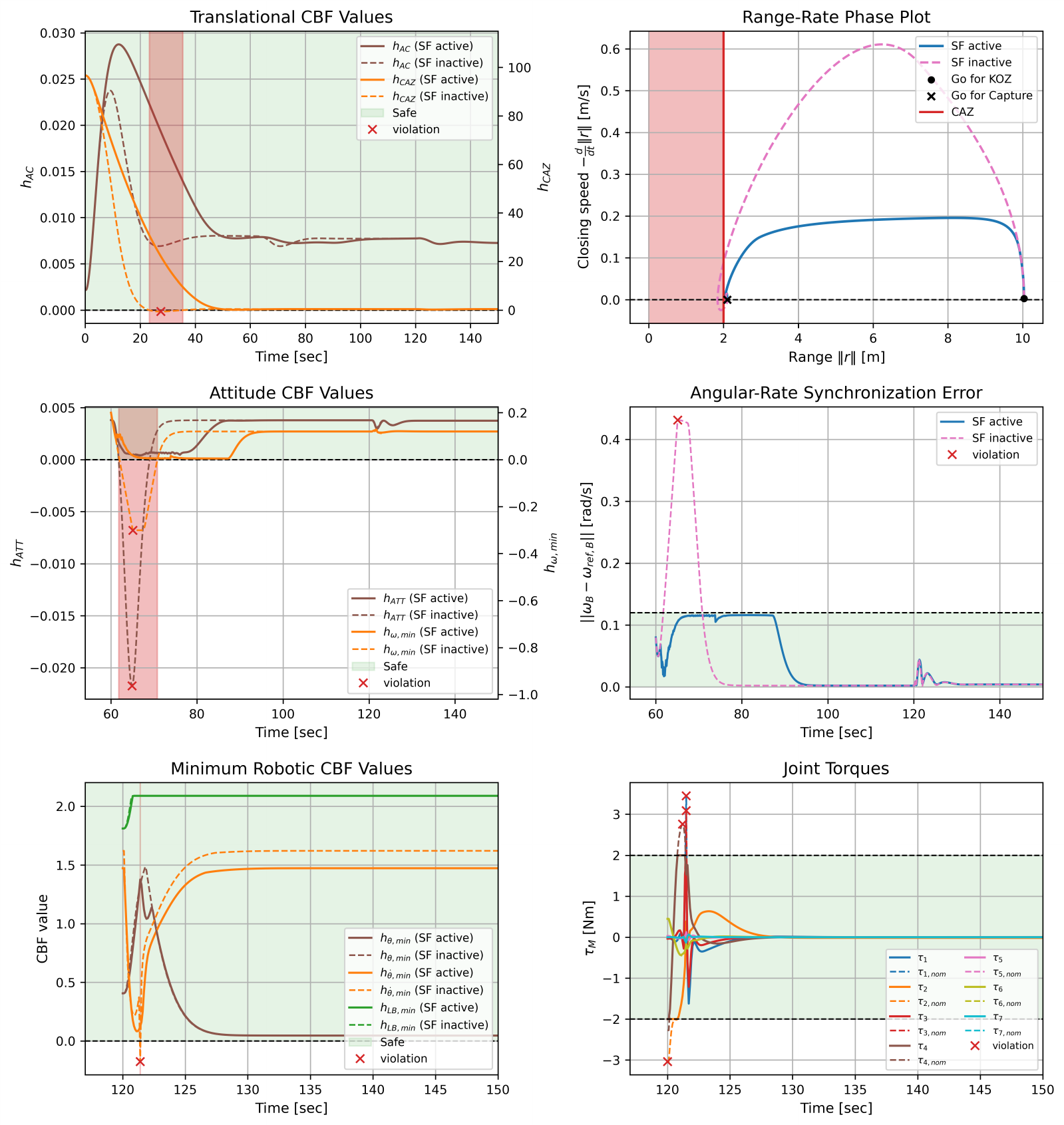}
	\caption{CBF evaluation demonstrating safety constraint satisfaction for the activated safety filters and violations for the nominal controllers highlighting the translational (top left), attitude (center left), and robotic subsystems (bottom left). Range-rate profile showing how the CBF-encoded operational zones influence the radial closing speed as the relative distance approaches the CAZ boundary (top right). Relative synchronization error with respect to the tumbling target rate (center right), and joint torque profile during the grasping maneuver (bottom right).}
	\label{fig:results_cbf}
\end{figure}
\subsection{Numerical Results}
The proposed control framework is validated in an OOS scenario using the astrodynamics simulator \textit{Basilisk} \cite{Kenneally2020}, where phases (II) and (III) start after $60\,\unit{s}$ and $120\,\unit{s}$, respectively. Bounds on the input and states are set to $\|\vc f_B\|_{\infty}\leq7\,\unit{N}$, $\|\vc{\tau}_B\|_{\infty}\leq10\,\unit{Nm}$, $\|\vc{\tau}_M\|_{\infty}\leq2\,\unit{Nm}$,
$\psi_{\mathrm{AC}}=\beta_{\mathrm{ATT}}=5\unit{\degree}$, $\|\vc\omega_B\|_{\infty}\leq0.2\,\unit{rad/s}$, along with CBF parameters $\gamma_h=1$, $\epsilon_{\mathrm{CAZ}}=0.01$, $\epsilon_{\mathrm{AC}}=2\times10^{-5}$, $\epsilon_{\mathrm{ATT}}=5\times10^{-4}$, $\epsilon_{\omega}=0.005$, $\epsilon_{\theta}=0.03$, $\epsilon_{\dot{\theta}}=0.01$, $\epsilon_{\mathrm{LB}}=0.03$, and equal sample time for all subsystems $T_s=0.1\unit{s}$, while the target tumbles with a constant rate $\vc\omega_{\mathrm{ref},B}=[-0.08,0,0]^\top\,\unit{rad/s}$ in a circular orbit at $600\,\unit{km}$ altitude.
The simulation results in Figure \ref{fig:results_trajectories} and \ref{fig:results_cbf} validate that the modular safety layer enforces the proposed CBF constraints, even when the unconstrained nominal controllers would lead to safety violations. In the translational subsystem, the safety filter preserves the conical AC and spherical CAZ, with the CBFs implicitly reducing the admissible radial closing speed as the servicer approaches the safety boundary. The attitude filter maintains LOS pointing and angular rate synchronization, while the robotic safety intervention is primarily driven by joint velocity and torque constraints rather than self-collision avoidance. Notably, constraint satisfaction is retained during the phase transitions and during simultaneous translational, attitude synchronization, and robotic motion, indicating that the sequential coupling-aware safety filters remain compatible in the considered multi-phase scenario.
\subsection{Monte Carlo Simulation}
A Monte Carlo simulation over 100 trajectories investigates the robustness of the proposed safety layer against parameter sensitivity in the initial position and the target tumbling rate. Starting at different sampled distances to the target within the AC, the target tumbling rate about its $x$-axis is varied uniformly from $0$ to $0.15\,\unit{rad/s}$.

\begin{figure}[b!]
		\vspace{-0.35cm}
	\centering
	\includegraphics[width=0.9\linewidth,page=5]{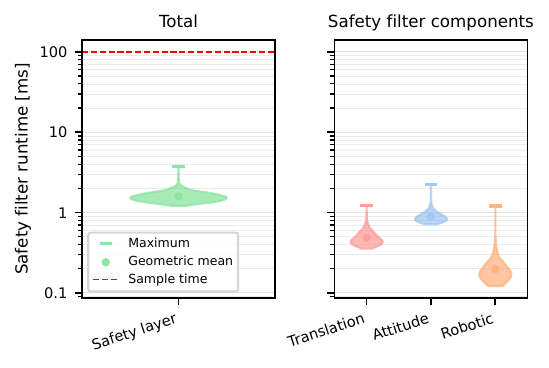}
	\caption{Computational runtime analysis over the entire Monte Carlo simulation. Maximum runtime and geometric mean are visualized for the combined safety layer (left) and for each subsystem safety filter QP (right).}
	\label{fig:computational-demand}
\end{figure}

In Figure \ref{fig:computational-demand}, the computational demand is illustrated for the combined safety layer and the respective subsystem QPs over the course of the Monte Carlo simulation. The geometric mean of the total runtime is $1.59\,\unit{ms}$, almost two orders of magnitude below the $100\,\unit{ms}$ controller sample time. Its maximum total solve time is $3.75\,\unit{ms}$. Although the attitude safety filter requires the highest computational effort compared to the translational and robotic QPs, all component runtimes remain below $2.3\,\unit{ms}$. The results demonstrate online applicability of the modular safety layer and emphasize its runtime-efficient implementation, using the CasADi \cite{Andersson2019} framework and OSQP \cite{Stellato2020} solver.

\begin{figure*}[h]
	\setlength{\belowcaptionskip}{-15pt}
	\centering
	\includegraphics[width=1.0\linewidth,page=6]{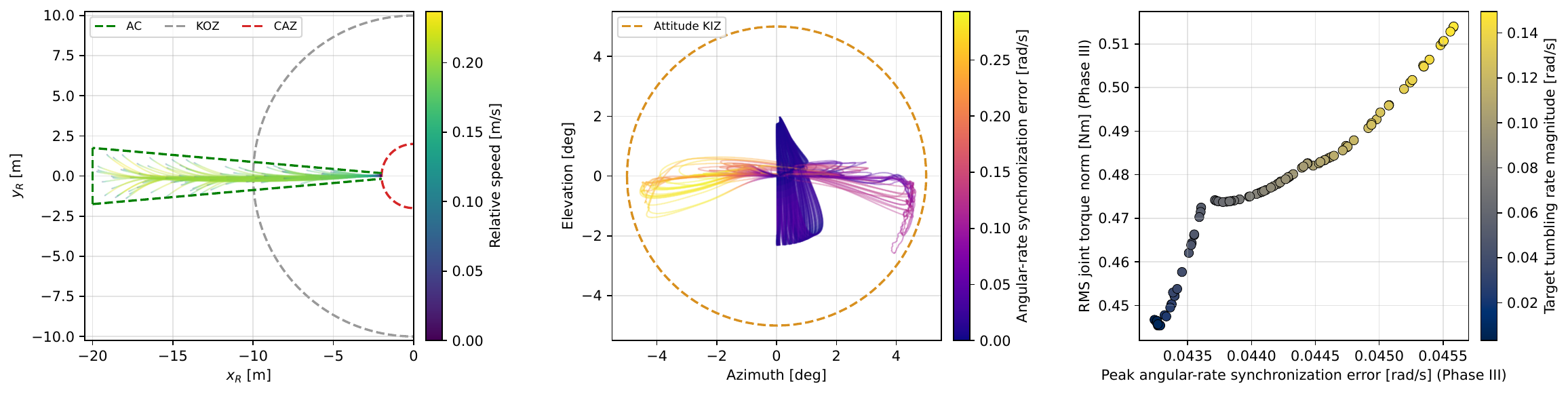}
	\caption{Monte Carlo simulation results. Left: Closed-loop translational approach trajectories projected onto the $x_R$-$y_R$-plane, colored by relative speed and satisfying the AC and CAZ operational constraints. Center: Closed-loop attitude motion profile, colored by the servicer angular rate synchronization error and within the bounds of the line-of-sight cone. Right: Relation between the robotic joint torque norm and the peak angular rate synchronization error during grasping in phase (III), illustrated per run and colored by the varying target tumbling rate.}
	\label{fig:monte-carlo}
\end{figure*}
Figure \ref{fig:monte-carlo} depicts the results of the Monte Carlo simulation over the entire mission life cycle, highlighting that each trajectory satisfies the narrow operational zones in both the translational and attitude state space. The scatter analysis provides deeper insights into the robustness against higher tumbling rates during the grasping maneuver in phase (III). Additionally, it is consistent with the expected backward dynamic coupling from the robotic motion to the servicer base. Increasing the scenario complexity through larger tumbling rates is associated with a higher robotic control effort and a moderately larger response in the angular rate synchronization error. However, the observed range in the peak synchronization error of approximately $0.0432$-$0.0457\,\unit{rad/s}$ during grasping (III) remains significantly below the larger synchronization errors of up to $0.3\,\unit{rad/s}$ during the preceding phase (II), as indicated in the center plot. Likewise, the low transient increase in synchronization error at the start of robotic grasping (III), coincident with manipulator motion, together with the less critical attitude CBF values compared to the angular rate synchronization phase (II) is also visualized in Figure \ref{fig:results_cbf}. While the backward coupling from the robotic arm to the servicer base is not explicitly compensated for in the proposed safety layer, the Monte Carlo simulation results demonstrate sufficient robustness even for the investigated higher tumbling rates.
\section{Conclusion} \label{sec: Conclusion}
In this paper, we developed a safe control framework for a robotic OOS mission that reflects modular subsystem-based flight algorithm design common in large space mission consortia. To this end, we incorporated extensive CBF constraints for a 13-DoF space robot following the requirements of latest CPO guidelines, thereby extending the CBF coverage of existing OOS approaches. The investigation of contact forces and formal safety guarantees during detumbling remain subject to future work.


\begin{thebibliography}{10}
\providecommand{\url}[1]{#1}
\csname url@samestyle\endcsname
\providecommand{\newblock}{\relax}
\providecommand{\bibinfo}[2]{#2}
\providecommand{\BIBentrySTDinterwordspacing}{\spaceskip=0pt\relax}
\providecommand{\BIBentryALTinterwordstretchfactor}{4}
\providecommand{\BIBentryALTinterwordspacing}{\spaceskip=\fontdimen2\font plus
\BIBentryALTinterwordstretchfactor\fontdimen3\font minus
  \fontdimen4\font\relax}
\providecommand{\BIBforeignlanguage}[2]{{%
\expandafter\ifx\csname l@#1\endcsname\relax
\typeout{** WARNING: IEEEtran.bst: No hyphenation pattern has been}%
\typeout{** loaded for the language `#1'. Using the pattern for}%
\typeout{** the default language instead.}%
\else
\language=\csname l@#1\endcsname
\fi
#2}}
\providecommand{\BIBdecl}{\relax}
\BIBdecl

\bibitem{ESA2024}
ESA, ``{ESA-TECSYE-TN-022522—Guidelines on Safe Close Proximity
  Operations},'' Tech. Rep., 2024.

\bibitem{ISO243302022}
ISO24330, ``{Space systems - Rendezvous and Proximity Operations (RPO) and On
  Orbit Servicing (OOS) - Programmatic principles and practices},'' 2022.

\bibitem{Vasconcelos2025}
J.~Vasconcelos, S.~Gaggi, T.~Amaral, C.~Bakouche, A.~Cotuna, and A.~Friaças,
  ``{Close-Proximity Operations Design, Analysis, and Validation for
  Non-Cooperative Targets with an Application to the ClearSpace-1 Mission},''
  \emph{Aerospace}, vol.~12, 1 2025.

\bibitem{Wabersich2023}
K.~P. Wabersich \emph{et~al.}, ``{Data-Driven Safety Filters: Hamilton-Jacobi
  Reachability, Control Barrier Functions, and Predictive Methods for Uncertain
  Systems},'' \emph{IEEE Control Systems}, vol.~43, pp. 137--177, 9 2023.

\bibitem{Ames2019}
A.~D. Ames, S.~Coogan, M.~Egerstedt, G.~Notomista, K.~Sreenath, and P.~Tabuada,
  ``{Control Barrier Functions: Theory and Applications},'' in \emph{2019 18th
  European Control Conference (ECC)}, 2019, pp. 3420--3431.

\bibitem{Breeden2022Docking}
J.~Breeden and D.~Panagou, ``{Guaranteed Safe Spacecraft Docking with Control
  Barrier Functions},'' \emph{IEEE Control Systems Letters}, vol.~6, pp.
  2000--2005, 2022.

\bibitem{Meinert2026}
A.~Meinert, N.~Baldauf, P.~Stadler, and A.~Turnwald, ``{Safety-Guaranteed
  Imitation Learning from Nonlinear Model Predictive Control for Spacecraft
  Close Proximity Operations},'' in \emph{2026 European Control Conference
  (ECC)}, 2026, pp. 1453--1458.

\bibitem{Tan2020}
X.~Tan and D.~V. Dimarogonas, ``{Construction of control barrier function and
  C2 reference trajectory for constrained attitude maneuvers},'' in \emph{2020
  59th IEEE Conference on Decision and Control (CDC)}, 2020, pp. 3329--3334.

\bibitem{vanWijk2024}
D.~E.~J. van Wijk, S.~Coogan, T.~G. Molnar, M.~Majji, and K.~L. Hobbs,
  ``{Disturbance-Robust Backup Control Barrier Functions: Safety Under
  Uncertain Dynamics},'' \emph{IEEE Control Systems Letters}, pp. 2817--2822,
  12 2024.

\bibitem{Morton2025}
D.~Morton and M.~Pavone, ``{Safe, Task-Consistent Manipulation with Operational
  Space Control Barrier Functions},'' in \emph{2025 IEEE/RSJ International
  Conference on Intelligent Robots and Systems (IROS)}, 3 2025, pp. 187--194.

\bibitem{Shen2024}
Z.~Shen, M.~Saveriano, F.~J. Abu-Dakka, and S.~Haddadin, ``{Safe Execution of
  Learned Orientation Skills with Conic Control Barrier Functions},'' in
  \emph{2024 IEEE International Conference on Robotics and Automation (ICRA)},
  3 2024, pp. 13\,376--13\,382.

\bibitem{Loettgen2025}
J.~L. Loettgen, K.~Worrall, G.~Aragon-Camerasa, M.~Ceriotti, and
  R.~Lampariello, ``{Control Barrier Functions for Safe Real-Time Control of
  Spacecraft Onboard Robotic Manipulators},'' in \emph{2025 International
  Conference on Space Robotics (iSpaRo)}, 2025, pp. 539--546.

\bibitem{Sharifi2024}
M.~Sharifi and S.~Heshmati-Alamdari, ``{Safe Force/Position Tracking Control
  via Control Barrier Functions for Floating Base Mobile Manipulator
  Systems},'' in \emph{2024 European Control Conference (ECC)}, 2024, pp.
  3650--3655.

\bibitem{Shi2025}
C.~Shi, T.~Meng, K.~Wang, J.~Lei, W.~Wang, and R.~Mao, ``{Safe tracking control
  for free-flying space robots via control barrier functions},'' \emph{Robotics
  and Autonomous Systems}, vol. 184, p. 104865, 2 2025.

\bibitem{Invernizzi2023}
D.~Invernizzi \emph{et~al.}, ``{Robust Control of Free-Flying Space Manipulator
  for Capturing Uncontrolled Tumbling Objects},'' in \emph{Proceedings of the
  12th International ESA Conference on Guidance, Navigation and Control Systems
  (GNC)}, 2023.

\bibitem{Colmenarejo2018}
P.~Colmenarejo \emph{et~al.}, ``{Methods and outcomes of the COMRADE project -
  Design of robust Combined control for robotic spacecraft and manipulator in
  servicing missions: comparison between Hinf and nonlinear Lyapunov-based
  approaches},'' in \emph{69th International Astronautical Congress (IAC)},
  2018.

\bibitem{Kenneally2020}
P.~W. Kenneally, S.~Piggott, and H.~Schaub, ``{Basilisk: A Flexible, Scalable
  and Modular Astrodynamics Simulation Framework},'' \emph{Journal of Aerospace
  Information Systems}, vol.~17, pp. 496--507, 2020.

\bibitem{Breeden2023Robust}
J.~Breeden and D.~Panagou, ``{Robust Control Barrier Functions under high
  relative degree and input constraints for satellite trajectories},''
  \emph{Automatica}, vol. 155, p. 111109, 9 2023.

\bibitem{Stadler2026}
P.~Stadler, A.~Meinert, N.~Baldauf, and A.~Turnwald, ``{Lightweight Model
  Predictive Control for Spacecraft Rendezvous Attitude Synchronization},'' in
  \emph{2026 European Control Conference (ECC)}, 2026, pp. 3748--3753.

\bibitem{Andersson2019}
J.~A.~E. Andersson, J.~Gillis, G.~Horn, J.~B. Rawlings, and M.~Diehl,
  ``{CasADi---A Software Framework for Nonlinear Optimization and Optimal
  Control},'' \emph{Mathematical Programming Computation}, pp. 1--36, 2019.

\bibitem{Stellato2020}
B.~Stellato, G.~Banjac, P.~Goulart, A.~Bemporad, and S.~Boyd, ``{OSQP: An
  Operator Splitting Solver for Quadratic Programs},'' \emph{Mathematical
  Programming Computation}, pp. 637--672, 2020.

\end{thebibliography}
\end{document}